\documentclass[preprint,12pt]{elsarticle}

\usepackage{subcaption}
\usepackage[T5,T1]{fontenc}
\usepackage{multirow}

\DeclareTextFontCommand{\textvietnamese}{\fontencoding{T5}\selectfont}

\usepackage{amssymb}
\usepackage{url}
\usepackage{amsmath}
\usepackage{booktabs}
\usepackage[hidelinks]{hyperref}
\usepackage{xcolor}
\hypersetup{%
        colorlinks,
        breaklinks=true,
        plainpages=false,%
        citecolor=blue,
        linkcolor=blue,
        urlcolor=blue,
        bookmarksopen=true,%
        bookmarksnumbered=false,%
        bookmarksdepth=5%
}

\usepackage{pifont}
\newcommand{\cmark}{\ding{51}}%
\newcommand{\xmark}{\ding{55}}%
\journal{Computers and Electrical Engineering}

\begin{document}

\begin{frontmatter}



\title{Advancing Vietnamese Visual Question Answering with Transformer and Convolutional Integration}


\author[1,3]{Ngoc Son Nguyen}

\author[1,3]{Son Van Nguyen}

\author[2,3]{Tung Le\corref{cor1}}
\ead{lttung@fit.hcmus.edu.vn}

\cortext[cor1]{Corresponding author}

\affiliation[1]{organization={Faculty of Mathematics and Computer Science, University of Science},
            city={\\Ho Chi Minh},
            country={Vietnam}}

\affiliation[2]{organization={Faculty of Information Technology, University of Science},
            city={Ho Chi Minh},
            country={Vietnam}}

\affiliation[3]{organization={Vietnam National University},
            city={Ho Chi Minh},
            country={Vietnam}}


\begin{abstract}
 Visual Question Answering (VQA) has recently emerged as a potential research domain, captivating the interest of many in the field of artificial intelligence and computer vision. Despite the prevalence of approaches in English, there is a notable lack of systems specifically developed for certain languages, particularly Vietnamese. This study aims to bridge this gap by conducting comprehensive experiments on the Vietnamese Visual Question Answering (ViVQA) dataset, demonstrating the effectiveness of our proposed model. In response to community interest, we have developed a model that enhances image representation capabilities, thereby improving overall performance in the ViVQA system. Specifically, our model integrates the Bootstrapping Language-Image Pre-training with frozen unimodal models (BLIP-2) and the convolutional neural network EfficientNet to extract and process both local and global features from images. This integration leverages the strengths of transformer-based architectures for capturing comprehensive contextual information and convolutional networks for detailed local features. By freezing the parameters of these pre-trained models, we significantly reduce the computational cost and training time, while maintaining high performance. This approach significantly improves image representation and enhances the performance of existing VQA systems. We then leverage a multi-modal fusion module based on a general-purpose multi-modal foundation model (BEiT-3) to fuse the information between visual and textual features. Our experimental findings demonstrate that our model surpasses competing baselines, achieving promising performance. This is particularly evident in its accuracy of $71.04\%$ on the test set of the ViVQA dataset, marking a significant advancement in our research area. The code is available at \url{https://github.com/nngocson2002/ViVQA}.
\end{abstract}






\end{frontmatter}


\section{Introduction}
\label{sec:Introduction}
The rapid enhancement of multi-modal processing in recent years~\cite{SHI2023108641, WANG2022108002, XING2022108270, LI2023108855} has not only opened up vast potential but has also considered a range of remarkable challenges. Among these, the field of VQA emerges as a critical problem in comprehending multi-modalities. This task entails empowering computers to interpret and respond to queries based on the content of images, necessitating a synergistic integration of advancements in both image processing and natural language processing. Despite its complexities, VQA is an essential and challenging task in the ongoing evolution of multimedia data. It represents a significant step towards understanding and integrating diverse forms of information, a cornerstone in the development of more intelligent and versatile Artificial Intelligence (AI) systems.



However, a notable challenge is that the majority of datasets used to train current VQA models are in English. Most research efforts in this field have prioritized languages with abundant resources, leaving low-resource languages like Vietnamese largely overlooked. This presents significant issues regarding the applicability of VQA models to questions written in Vietnamese and their ability to comprehend Vietnamese culture. In this context, focusing on the ViVQA task is invaluable, as it plays an important role in driving the development of innovative approaches in this specific domain. To address this gap, an increasing number of VQA models for Vietnamese have been established, such as those proposed in recent studies~\cite{tran-etal-2021-vivqa, nguyen-tran-etal-2022-bi, pat, bartphobeit}, significantly contributing to this field. Our work is also no exception, aiming to tackle the ViVQA challenge by developing a novel system for ViVQA that leverages models based on the Transformer architecture~\cite{vaswani2017attention}, which have been pre-trained on various tasks related to both vision and language, such as UNITER ~\cite{uniter}, CLIP~\cite{clip}, BLIP~\cite{li2022blip}, and Flamingo~\cite{alayrac2022flamingo}.

A fundamental aspect of VQA systems is the effective extraction and representation of textual and visual features in meaningful ways. The development of pre-trained models in text and image processing plays a crucial role in this aspect. However, it is essential to recognize that each type of model brings unique strengths to the table and should be evaluated from diverse perspectives. In our model, multi-modal features are intricately connected and integrated through a specialized fusion module. Building on the successes of previous models, our approach focuses on refining the semantic representations of uni-modal.

Although the Transformer architecture is renowned for its ability to analyze complete images, it sometimes overlooks crucial small details. This oversight is a significant challenge in VQA, as these details often contain vital information for understanding and answering questions. Shen et al.~\cite{shen2023} contended that a VQA model relied on the Transformer architecture, exclusively considering global dependencies, is unable to effectively capture image context information. Through studies, they also proved that using both local and global features in image processing improves performance. Clearly, it is essential to consider the contextual representation of images from both local and global perspectives. Therefore, with this consideration in mind, we propose an innovative integration of transformer-based and convolutional methods in image processing to enhance the feature extraction module in VQA. This integration aims to enhance the visual information by combining the detailed local features extracted by the Convolutional Neural Network (CNN) with the broad global features derived from Vision Transformer-based models. Our novel component has the potential to evolve alongside advancements in visual feature extraction models by incorporating current improvements into our sophisticated combination. The promising results on the Vietnamese VQA dataset demonstrate that our proposed approach is more efficient than existing competitive baselines and plays an important role in representing images in VQA processing.

This paper makes several pivotal contributions to the field of ViVQA, highlighted as follows:
\begin{itemize}
    \item We propose an innovative methodological advancement by integrating transformer-based and convolutional methods in image processing. Our novel representation consists of local and global consideration through the content of images.
    \item Our novel visual feature extraction is integrated into the ViVQA approach, employing the latest fusion module to deploy the efficient system.
    \item Through the detailed experimental analysis and comparative results, our  method demonstrates the potential performance over existing competitive baselines. This achievement is a significant stride in leveraging multi-view feature representation, enhancing the understanding of different modalities within ViVQA. 
\end{itemize}
\section{Related Works}\label{sec:RW}
\subsection{Overview}
VQA methods typically consist of three primary components: modules for feature extraction from textual and visual inputs, a multi-modal fusion module, and a classifier or generator module. Initially, Antol et al.~\cite{antol2015vqa} introduced the VQA task alongside a simple baseline, utilizing VGGNet~\cite{vgg} and Long Short-Term Memory (LSTM)~\cite{lstm} for visual and textual feature extraction, respectively. They then fused these features into a joint representation for classifying candidate answers. Subsequently, Yuling Xi et al.~\cite{yulingxi} introduced an approach focusing on object relationships within images to address VQA tasks. Their method utilized a question-based attention mechanism to highlight relevant object regions, enabling accurate responses to questions regarding image content. Additionally, Yirui Wu et al.~\cite{yiruiwu} proposed a baseline method emphasizing both global and local object relationships, underscoring the significance of comprehensive relation reasoning. Their approach leveraged information from the entire image for global understanding while the local aspect focused on modeling relationships among multiple objects to facilitate answer generation. Notably, these baselines all share the commonality of employing pre-trained CNN models like ResNet~\cite{resnet}, VGGNet~\cite{vgg}, and FasterRCNN~\cite{ren2015faster} for image feature extraction. In terms of text embedding, pre-trained word embeddings such as GloVe~\cite{glove} or Word2Vec~\cite{word2vec} are often used along with LSTM~\cite{lstm} or Gated Recurrent Unit (GRU)~\cite{gru} networks to extract linguistic features. In recent years, researchers have increasingly turned to Transformer-based models such as ViLBERT~\cite{lu2019vilbert}, X-LXMERT~\cite{x-lxmert}, and VLMo~\cite{bao2022vlmo}. Consequently, later works often rely on these models, leveraging the capabilities of pre-trained models.

The primary focus of this paper is the examination of monolingual ViVQA. In previous research, Tran et al.~\cite{tran-etal-2021-vivqa} introduced a model that utilized Hierarchical Co-Attention~\cite{lu2016hierarchical}, establishing the first baseline for ViVQA dataset. Their experiments demonstrated that their system outperformed two baseline models, LSTM~\cite{lstm} and BiLSTM, on the ViQA dataset. Following this, with the advent of transformers, Nguyen-Tran et al.~\cite{nguyen-tran-etal-2022-bi} proposed an approach that incorporates a Bi-directional Cross-attention architecture. This model was designed to more effectively comprehend the complex interplay between visual and linguistic features, specifically in the Vietnamese context. Another notable contribution by Nguyen et al.~\cite{pat} was the introduction of Parallel Attention Transformer (PAT), a novel multimodal learning scheme. This method effectively integrates the advantages of grammar and contextual understanding in Vietnamese. More recently, Tran et al.~\cite{bartphobeit} proposed leveraging BEiT-3~\cite{Wang_2023_CVPR} as a fusion encoder to intricately model the deep interactions between images and questions. Furthermore, they employed BARTpho~\cite{bartpho}, a monolingual sequence-to-sequence model pre-trained for Vietnamese, to enhance question representations. Remarkably, this model has achieved the current state-of-the-art performance on the ViVQA dataset. This research significantly advances the field of VQA, particularly in the Vietnamese language context, showcasing the effectiveness of novel models in capturing the nuanced interactions between visual and linguistic elements.

\subsection{ViVQA Models in Previous Studies}
In this section, we will briefly present two methods that currently achieve the best performance on the ViVQA task and point out their weaknesses, which motivates our study.
\subsubsection{Multimodal Contextual Attention Model}
In $2022$, Anh Duc Nguyen et al.~\cite{CMSA} introduced the Multi-vision Contextual Attention (MCA) model, presenting an innovative approach to image feature extraction that seamlessly integrates both global and local features. This model excels in learning attentional embeddings, capturing diverse contexts from both images and questions while being mutually guided.

\begin{figure}[h!]
    \centering
    \includegraphics[width = \textwidth]{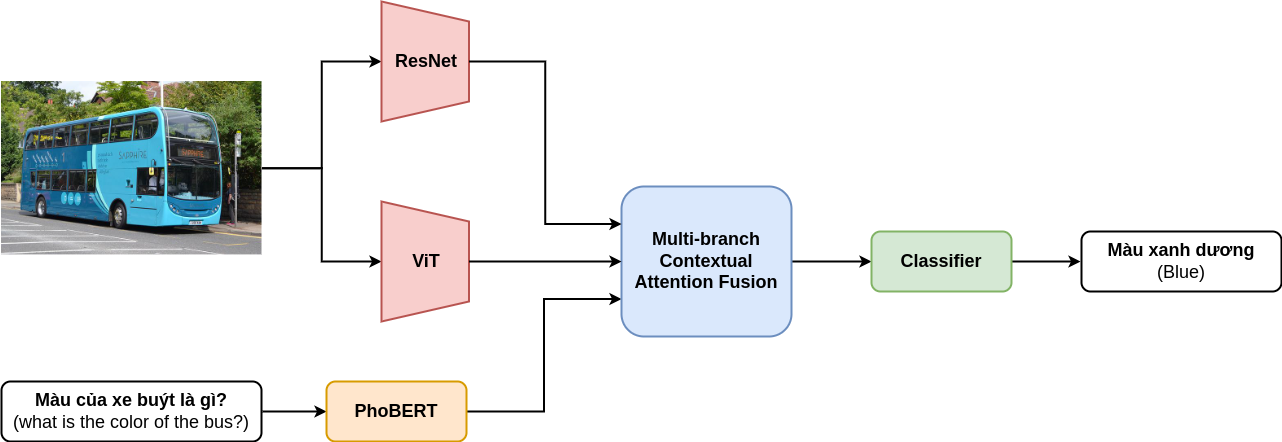}
    \caption{Architecture of Multi-vision Contextual Attention, which consists of four main modules: the features extraction networks to learn the image (\textbf{the red-colored shapes}) and question (\textbf{the orange-colored shape}) understanding; the multi-branch contextual attention fusion (\textbf{the blue-colored shape}) to fuse visual and textual features; and the classifier (\textbf{the green-colored shape}) to predict the answer.}
    \label{fig:mca}
\end{figure}

As depicted in Figure \ref{fig:mca}, the image initially extracts local features corresponding to specific regions or parts of objects in the image, acquired through the utilization of ResNet~\cite{resnet}, a convolutional neural network. Simultaneously, global features encapsulate fundamental information about the contextual surroundings within the image, including associations among objects, and are derived using the Vision Transformer (ViT)~\cite{vit}. For question embedding, PhoBERT~\cite{nguyen-tuan-nguyen-2020-phobert} is employed to extract features from questions. Based on the obtained visual and textual features, a multi-branch contextual attention fusion hierarchically integrates visual and textual attention features, guiding each other and fusing them into a multimodal feature representation. This pioneering model represents a significant advancement in feature extraction, showcasing the integration of diverse contextual elements for a more comprehensive understanding of multimodal data.

Despite the combination of CNN and a Transformer in this model, it's worth noting that ResNet is not the latest CNN model. In our proposed architecture, we aim to innovate by combining both local and global features inspired by the MCA model. Through this approach, we anticipate achieving better performance.
\subsubsection{BARTPhoBEiT}
Tran et al.~\cite{bartphobeit} have recently introduced an innovative methodology that integrates pre-trained sequence-to-sequence and image Transformers customized for the Vietnamese language. This pioneering approach draws inspiration from BEiT-3~\cite{Wang_2023_CVPR}, serving as the fundamental framework for their model.

\begin{figure}[h!]
    \centering
    \includegraphics[width = \textwidth]{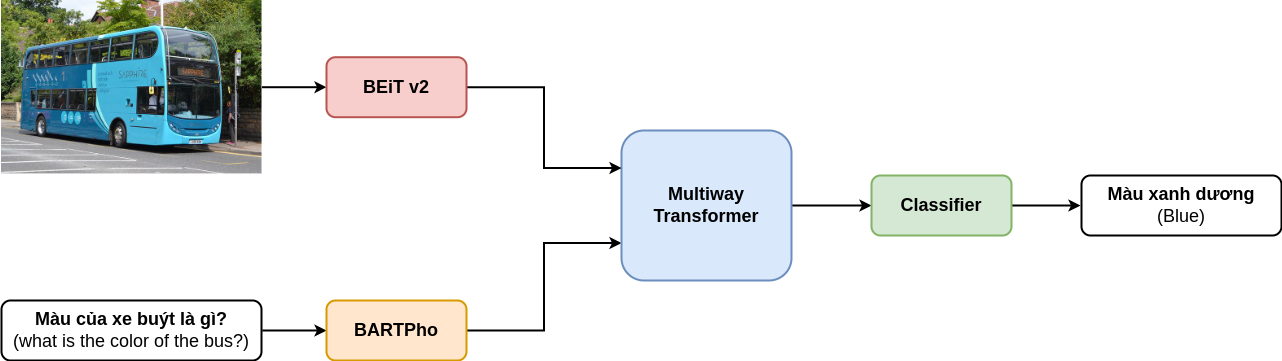}
    \caption{The architecture of the BARTPhoBEiT model, which comprise three main modules: the image embedding (\textbf{the red-colored shape}), the question embedding (\textbf{the orange-colored shape}); the multimodal fusion (\textbf{the blue-colored shape}); and the classifier (\textbf{the green-colored shape}) to predict the answer.}
    \label{fig:bartphobeit}
\end{figure}

As shown in Figure \ref{fig:bartphobeit}, the image embedding module utilizes BEiT v2~\cite{peng2022unified} to extract global features from images, while for question embedding, BARTPho~\cite{bartpho} is employed to extract question features. Subsequently, the visual and question features are fused using a Multiway Transformer architecture~\cite{bao2022vlmo}. These Transformers dynamically route information to the appropriate expert for each modality, ensuring the capture of modality-specific details. This unique feature significantly enhances the model's efficacy in handling diverse data modalities, emphasizing its versatility in cross-modal applications. Furthermore, the robustness of this model relies on the Transformer architecture, enabling seamless integration of an extensive volume of pre-trained data across diverse domains, resulting in state-of-the-art performance on the ViVQA dataset.

Despite the commendable results of their work, we have identified the complexity of image processing and the absence of information about local features as areas for further exploration. Building on the insights from their study, we are motivated to focus on the development of methods that enhance the performance of image processing, particularly for the ViVQA task.

\subsection{Dataset for ViVQA}
In the context of Vietnamese, the development of data for VQA is not as advanced as it is in English. The available resources are limited, presenting a scarcity of data specifically tailored for ViVQA. This dearth underscores the challenges and gaps in the field, as only a modest amount of data is currently accessible for research and improvement in ViVQA. It is challenging to create models that can effectively generalize to new inputs when there is insufficient data to capture the entire variety of text and images. Furthermore, a significant element influencing the models' performance is the dataset's quality. Predictions might be erroneous and ineffective generalization can result from incomplete and low-quality data. Within the scope of this study, our primary focus is on monolingual Vietnamese datasets. We will survey datasets that align with our research objectives to serve as the foundation for our model.
\subsubsection{OpenViVQA Dataset}
In $2023$, Nghia et al. \cite{Nguyen_2023} introduced the OpenViVQA dataset~\cite{open-vivqadataset} aims to showcase the inherently open-ended nature of both questions and answers. In order to seamlessly align with the intricacies of the Vietnamese language, the OpenViVQA dataset thoughtfully incorporates a diverse array of images captured within the rich tapestry of Vietnam's landscapes. This deliberate inclusion not only serves to showcase the linguistic nuances but also aims to highlight the myriad characteristics that distinctly define Vietnam in comparison to other regions. By strategically curating scenes from Vietnam, the dataset offers a comprehensive representation of the country's unique cultural, environmental, and visual elements, contributing to a nuanced understanding within the realm of VQA research.

The dataset is divided into two distinct components: $``$Text QA$"$ and $``$Non-text QA$"$. Specifically, the non-text QAs are designed to extract information pertinent to objects, encompassing their attributes and relational aspects. On the contrary, it is categorized as Text QA if it effectively leverages textual information emanating from the images themselves or employs the words present within the images as a means to pose questions. As a result of the inclusion of Text QA within the dataset, consequently making Optical Character Recognition (OCR)~\cite{ocr-vqa} a more involved process, our primary focus does not lie in this domain. Our paramount focus is intricately tied to the detailed examination of large objects depicted in the images. Thus, the OpenViVQA dataset does not align with our preference for this particular study.
\subsubsection{ViVQA Dataset} \label{vivqadataset}
The inaugural dataset in ViVQA, crafted by Tran et al.~\cite{tran-etal-2021-vivqa} in $2021$, stands as a pioneering contribution to the field. Drawing inspiration from established English VQA benchmarks and COCO-QA \cite{mscoco}, this dataset has evolved into a widely embraced benchmark for related research endeavors. The creation process involved leveraging machine translation for the conversion of questions and answers, followed by meticulous manual validation to ensure correctness and linguistic fluency in the translated content. 

The ViVQA dataset~\cite{vivqadataset} encompasses a collection of $10,328$ images paired with $15,000$ corresponding questions, intricately tailored to reflect the content within the images. To facilitate experimentation and evaluation, the dataset underwent a random partitioning into training and test sets, maintaining an $80:20$ ratio. Noteworthy is the dataset's categorization of questions into four distinct types: Object, Number, Color, and Location, constituting $41.55\%$, $14.81\%$, $20.82\%$, and $22.82\%$ of the total questions, respectively. 

This dataset is particularly pertinent to our research objectives as our paper predominantly delves into advanced image processing techniques. Our primary focus lies in extracting detailed information related to objects, encompassing their attributes, relational aspects, and the overall landscape depicted across the image. Consequently, we have deliberately chosen this dataset for our study, opting against OpenViVQA. This decision is attributed to the existence of questions related to textual content within images in OpenViVQA, which does not align with the specific aspects of image analysis and processing we are exploring in our research.
\section{Methodology}\label{sec:Method}
In this paper, we approach the ViVQA task as a classification problem. Specifically, given an image $I$ and a question $Q$, the goal is to determine the most probable answer $\hat{a}$ from a predefined set of answer options $A$:
\begin{equation}
    \hat{a} = \arg\max_{a \in A} P(a|I,Q,A) \label{eq1}
\end{equation}

\begin{figure}[h!]
    \centering
    \includegraphics[width = \textwidth]{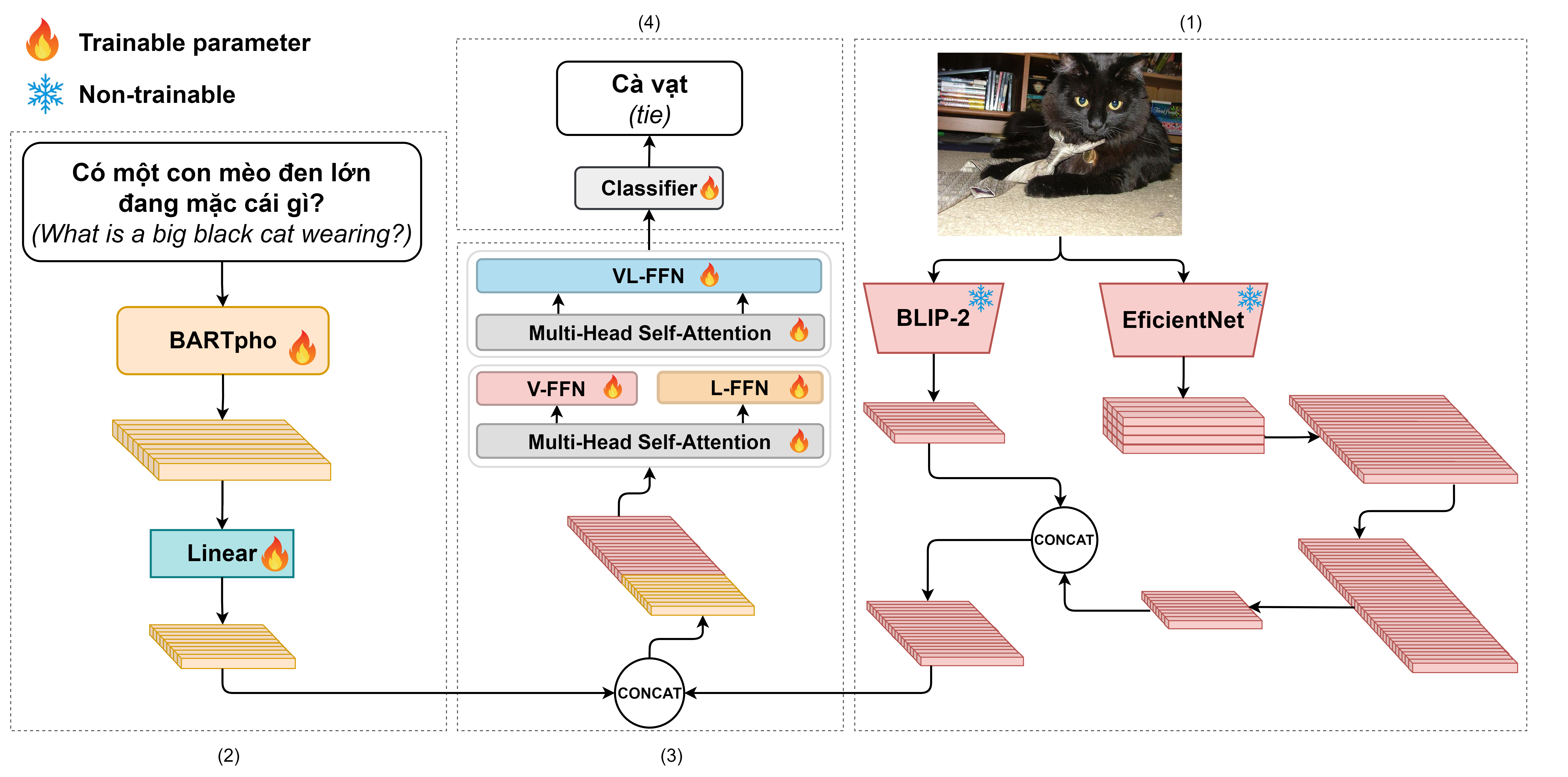}
    \caption{Overview of our architecture, which consists of four main components: (1) the Image Embedding module, (2) the Question Embedding module, (3) the Multi-modal Fusion module, and (4) the Classifier module for predicting the answer.}
    \label{fig:ourmodelfull}
\end{figure}

In this section, we present the detail of our ViVQA model architecture, illustrated in Figure \ref{fig:ourmodelfull}, drawing inspiration from BARTPhoBEiT~\cite{bartphobeit} with specific modifications in the visual processing aspect. We also provide detailed descriptions of the main image processing steps employed in our model. Our proposal advocates for the implementation of visual processing with frozen parameters, effectively reducing both the model parameters and training costs.

\subsection{Image Embedding Module} \label{imageembedding}
Utilizing the capabilities of pre-trained models, our image embedding module proves to be both effective and powerful, capturing valuable features through pretraining tasks. We employ both BLIP-2~\cite{li2023blip} and EfficientNet~\cite{pmlr-v97-tan19a} to extract visual features from both local and global respectively, amalgamating them to derive the final representation for the image. A distinctive feature of our module is its separation from the overall architecture, ensuring that its parameters remain static throughout the training process. This design choice significantly reduces the overall number of parameters in our model and shortens the training duration.

\subsubsection{BLIP-2}
The first model in the BLIP family, Bootstrapping Language-Image Pre-training for Unified Vision-Language Understanding and Generation (BLIP), was proposed by Li et al.~\cite{li2022blip}. BLIP introduced a novel Vision-Language Pre-training (VLP) framework that supports a wider range of downstream tasks than existing methods. This is achieved by jointly training models on extensive language and image datasets, thereby leveraging the complementary strengths of both modalities. 

Following this, Li et al.~\cite{li2023blip} proposed BLIP-2, an enhanced version of the original model with several adjustments in both the training strategy and the architecture. BLIP-2 introduces a pioneering method that leverages the capabilities of pre-trained vision models and language models. Unlike BLIP, which involves end-to-end training and often incurs high computational costs and lengthy training durations, BLIP-2 opts to freeze these pre-trained models during the training process. This strategy results in improved performance at a reduced computational cost and addresses the issue of catastrophic forgetting. Moreover, it ensures that the model capitalizes on the rich representations learned by the visual and language model, thereby enhancing its overall performance and effectiveness in comprehending both image and text inputs. With these strengths of BLIP-2, we decided to adopt it to take advantage of global image-level feature extraction.

\begin{figure}[h!]
    \centering
    \includegraphics[width = \textwidth]{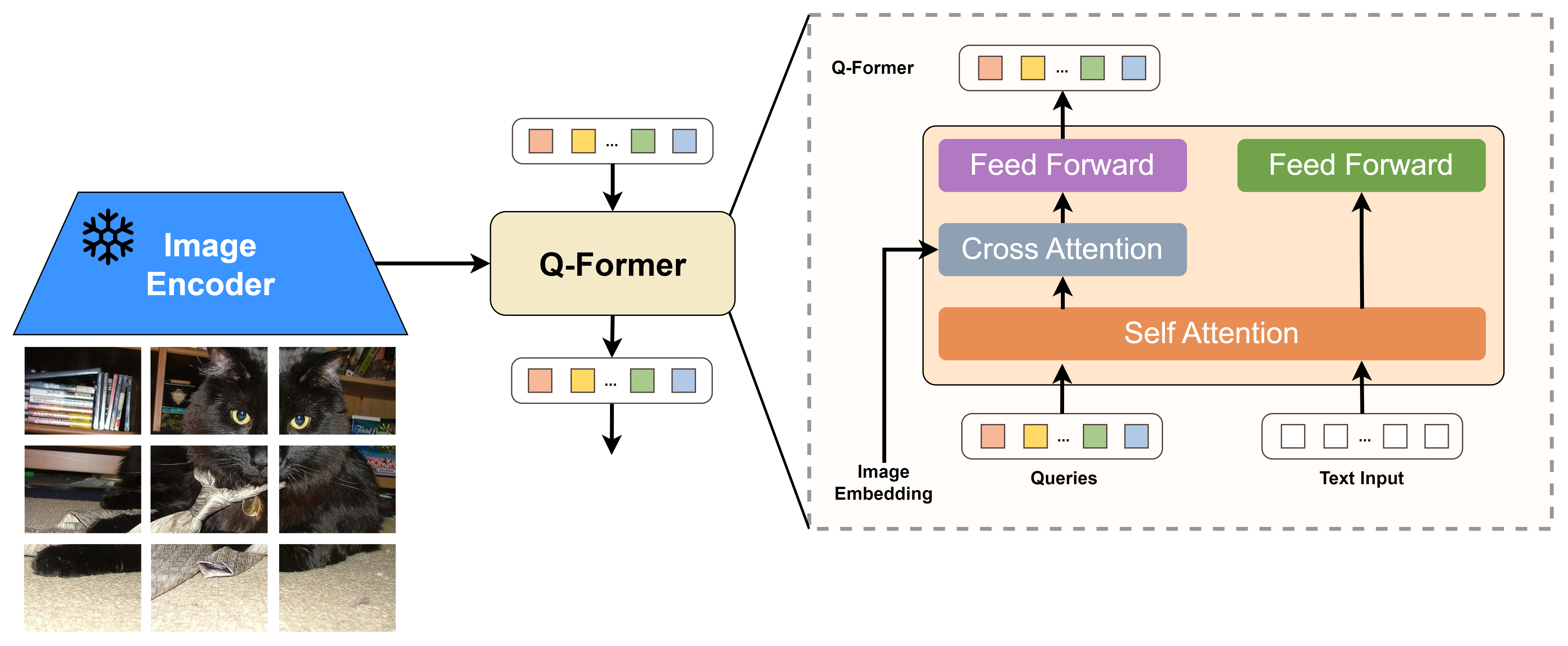}
    \caption{The model architecture of Q-Former in the BLIP-2 framework. We utilize the pre-trained Q-Former to extract visual features from the output embeddings of the frozen image encoder.}
    \label{fig:blip2}
\end{figure}

The central aim of BLIP-2 is to facilitate mutual understanding between the image and language. To achieve this objective, the Querying Transformer (Q-Former) module is specifically engineered to bridge the gap between vision and language. Following the Figure \ref{fig:blip2}, the Q-Former architecture in BLIP-2 incorporates (\textbf{left}) an image transformer utilizing an encoder-only transformer and (\textbf{right}) a text transformer, capable of functioning as both a text encoder and a text decoder. Both the image transformer and text transformer share the same self-attention layers. Given our focus on obtaining image representations for the ViVQA system, we are particularly interested in the image transformer within the Q-Former, which returns the visual features. A specific number of learnable query embeddings serve as input for the image transformer. The queries, each comprising 32 learned query tokens with a dimension of 768, contextualize the query tokens through the cross-attention mechanism with the representations of the image patches encoded by a large (frozen) ViT~\cite{vit}. The visual tokens serve as the output of the Q-Former.

Notably, a crucial stage involves representation learning, with the primary aim of pre-training within BLIP-2 being to refine the Q-Former's ability to enable queries to extract visual representations that are maximally informative regarding the accompanying text. Consequently, the features of the image generated by BLIP-2 encompass an ample amount of information, particularly in terms of global aspects, satisfying our ViVQA system's needs. The output query representation in $\mathbb{R}^{32 \times\ 768}$, significantly smaller than the size of the frozen image features (e.g., 257 × 1024 for ViT-L/14). Thus, the image $I$ representation after BLIP-2 feature extraction can be formulated as:
\begin{equation}
    V_G = \text{BLIP-2}(I) \in \mathbb{R}^{32 \times\ 768} \label{eq3}
\end{equation}

\subsubsection{EfficientNet}
We integrate the EfficientNet architecture introduced by Tan et al.~\cite{pmlr-v97-tan19a}, a sophisticated CNN, for local image-level feature extraction. Intuitively, CNN architectures inherently utilize kernels sliding over regions of the image, effectively capturing information locally. Previous CNN architectures commonly face a trade-off between model complexity and performance, wherein as performance improves, model complexity increases. This trade-off involves scaling, which can occur through depth scaling, width scaling, and resolution scaling.

Depth scaling entails increasing the depth of a neural network, typically resulting in improved performance. Width scaling involves increasing the number of channels (or neurons) in each layer, thereby enhancing the model's capacity to capture a broader range of features at each level and improving its representational power. Resolution scaling, meanwhile, requires resizing input images to higher resolutions, potentially leading to better performance as higher-resolution images contain more detailed information. 

EfficientNet~\cite{pmlr-v97-tan19a} is an improvement version basing on CNN that implements the concept of compound scaling, which enhances performance by uniformly scaling all dimensions of depth, width, and resolution. Its power is in how these three dimensions are balanced and adjusted using a systematic methodology. The model employs Mobile Inverted Bottleneck (MBConv)~\cite{sandler2018mobilenetv2, tan2019mnasnet} layers, combining depth-wise separable convolutions and inverted residual blocks. This bottleneck design enables efficient learning while preserving a high degree of representational power. The model's potency has been demonstrated through experiments by Tan et al.~\cite{pmlr-v97-tan19a}, which showed that models in the EfficientNet family achieve performance equivalent to other models while maintaining low complexity. To leverage these capabilities, we employ an EfficientNet-B7 model, a pre-trained model based on adversarial training~\cite{Xie_2020_CVPR}. It is initialized with pre-trained weights on ImageNet~\cite{image-net}, providing a robust foundation for directly extracting features for our ViVQA system.

The input image $I$, denoted as $\mathbb{R}^{C \times H \times\ W}$, where $C,H$ and $W$ stand for channels, height, and width respectively, undergoes processing within the pre-trained EfficientNet model. The features are extracted from the output of the final head before undergoing average pooling and a fully connected layer. This process generates a representation capturing the local features of the image, expressed as $\mathbb{R}^{2560 \times \frac{H}{32} \times\ \frac{W}{32}}$. It is noteworthy that our original image data is standardized to the dimensions of $\mathbb{R}^{3 \times 224 \times 224}$. Consequently, the image representation following EfficientNet feature extraction has dimensions $\mathbb{R}^{2560 \times 7 \times 7}$.
\begin{equation}
    V_L = \text{EfficientNet}(I) \in \mathbb{R}^{2560 \times 7 \times 7} \label{eq4}
\end{equation}

This approach allows us to capture intricate local features from the images, providing a rich and informative representation essential for subsequent stages in our model architecture. 

\subsubsection{Image Embedding}
In this stage, we will amalgamate the representations of BLIP-2 and EfficientNet to derive the final output of the image embedding module. Initially, an inconsistency was observed in the dimensions of BLIP-2 \eqref{eq3} and EfficientNet \eqref{eq4}. Thus, we will implement several steps to transform the representation of EfficientNet to align it with the output of BLIP-2. This transformation is crucial to facilitate operations for seamlessly combining the local and global features.

The transformation involves employing Adaptive Average Pooling to downsample from $\mathbb{R}^{2560 \times 7 \times 7}$ to $\mathbb{R}^{2560 \times 1 \times 32}$:
\begin{equation}
    V_L = \text{AdaptiveAveragePooling}(V_L) \in \mathbb{R}^{2560 \times 1 \times 32} \label{eq5}
\end{equation}

Subsequently, its dimensions are permuted to $\mathbb{R}^{32 \times 1 \times 2560}$, and another round of Adaptive Average Pooling is applied for downsampling to $\mathbb{R}^{32 \times 1 \times 768}$:
\begin{equation}
    V_L = \text{Permute}(V_L) \in \mathbb{R}^{32 \times 1 \times 2560} \label{eq6}
\end{equation}
\begin{equation}
    V_L = \text{AdaptiveAveragePooling}(V_L) \in \mathbb{R}^{32 \times 1 \times 768} \label{eq7}
\end{equation}

Through the flattening process, we derive the ultimate output of the image embedding module:
\begin{equation}
    V_L = \text{Flatten}(V_L) \in \mathbb{R}^{32 \times\ 768} \label{eq8}
\end{equation}

Finally, the fusion of local features and global features is computed as follows:

\begin{equation}
    V = \mathcal{F}(V_G, V_L) \in \mathbb{R}^{k \times 768} \label{eq9}
\end{equation}
where \(k\) relies on the operation $\mathcal{F}$. This equation represents the calculation for combining the global features, denoted as \(V_G\), and the local features, denoted as \(V_L\), through a fusion function $\mathcal{F}$.

\subsection{Text Embedding Module} \label{textembedding}

Throughout Tran et al.'s study~\cite{bartphobeit}, they observed the prevalence of grammatical errors and translation inaccuracies in ViVQA questions, even when validated by highly qualified annotators. Therefore, instead of relying on PhoBERT~\cite{nguyen-tuan-nguyen-2020-phobert} to extract question information, which predominantly comprehends word meanings in diverse contexts, our preference is to employ BARTpho~\cite{bartpho}. BARTpho stands out as a denoising autoencoder specifically crafted for pretraining sequence-to-sequence models geared towards Vietnamese language processing. The training regimen involves a two-step process: (1) corrupting the input text using a designated noising function and (2) learning to reconstruct the original text. This pretraining task empowers BARTpho with the capability to acquire robust representations resilient to noise, ambiguity, and variations in the input text. Hence, our approach leans towards leveraging the capabilities of BARTpho.

BARTpho offers two distinct variants, specifically BARTpho$_{syllyble}$, which is optimized for processing at the syllable level, and BARTpho$_{word}$, which is designed for handling language at the word level. In the case of BARTPho$_{word}$, the sentence must undergo preprocessing, being appropriately split into individual words and phrases. Subsequently, the input of complete sentences is fed into both the encoder and decoder, with the top hidden state of the decoder serving as a representation for each word. In our model, we use BARTpho$_{word}$ and the textual features $T$ can be articulated as:
\begin{equation}
    Q = \text{BARTpho}_{word}(T) \in \mathbb{R}^{(L+2) \times\ 1024} \label{eq2}
\end{equation}
where L + 2 with L is the max sequence length, adding 2 special token [$CLS$] and [$SEP$] representing the start-of-sentence and the end-of-sentence token.

Since the BLIP-2 in the image embedding module is predominantly trained on English data (specifically, images with English captions) and the text embedding module is tailored for Vietnamese, we set the parameters of the text embedding module to be trainable. This adaptability enables the text representation to dynamically adjust and align with the visual representation within the image embedding module.


\subsection{Multi-modal Fusion Module}
Before forwarding the representations of visual and question features to BEiT-3 for multimodal fusion, a Feed-Forward Network (FFN) layer is applied on the textual features. The purpose is to diminish the dimensionality from 1024 \eqref{eq2} to 768. This reduction is imperative due to the requirement of concatenating the embeddings of a given question and an image within this component. The dimension of the question is now given by:
\begin{equation}
    Q = \text{FFN}(Q) \in \mathbb{R}^{(L+2) \times\ 768} \label{eq10}
\end{equation}

This preprocessing step optimizes the dimensionality of the textual features, enhancing their compatibility for subsequent fusion with visual features. Subsequently, the visual \eqref{eq9} and question \eqref{eq10} features are concatenated:
\begin{equation}
    f_{VQ} = \text{concatenate}(V, Q) \in \mathbb{R}^{(k+L+2) \times 768} \label{eq11}
\end{equation}

Following this, the concatenated features are inputted into Multiway Transformers~\cite{bao2022vlmo} to jointly encode the image-question pair. Each Multiway Transformer block is composed of a shared self-attention module and a set of feed-forward networks tailored for diverse modalities. Each layer includes both a vision expert and a language expert, each responsible for processing the input tokens directed to them. Leveraging a group of modality experts incentivizes the model to capture a broader range of modality-specific information. The shared self-attention module acquires the ability to discern alignments between diverse modalities, facilitating profound fusion for tasks involving multiple modalities, such as vision-language tasks.

The result will have the same dimension as the input \(f_{VQ}\) \eqref{eq11}. The first vector in the output fusion is earmarked for classification, denoted as $\hat{f}_{VQ} \in \mathbb{R}^{1 \times 768}$.
Prior to being input into the Classifier module, \(\hat{f}_{VQ}\) undergoes processing through a pooler layer, involving the normalization, a FFN layer with hidden and output dimension of 768, and application of the hyperbolic tangent function as depicted below:
\begin{equation}
    \hat{f}_{VQ} = \text{Norm}(\hat{f}_{VQ}) \label{eq12}
\end{equation}
\begin{equation}
    \hat{f}_{VQ} = \text{FFN}(\hat{f}_{VQ}) \label{eq13}
\end{equation}
\begin{equation}
    \hat{f}_{VQ} = \text{Tanh}(\hat{f}_{VQ}) \label{eq14}
\end{equation}

\subsection{Classifier Module}
We integrate a classification head to predict the answer, consisting of the following components: (1) a FFN layer with hidden and output dimensions of 768 and 1536, (2) a normalization layer, (3) the GELU activation function, and (4) an additional FFN layer that maps the input dimension of 1536 to the number of classes (353 candidate answers).
\begin{equation}
    l = \text{FFN}(\hat{f}_{VQ})
\end{equation}
\begin{equation}
    l = \text{Norm}(l)
\end{equation}
\begin{equation}
    l = \text{GELU}(l)
\end{equation}
\begin{equation}
    l = \text{FFN}(l)
\end{equation}

Following this, a softmax function is applied to convert the output into a distribution within the range of 0 to 1:
\begin{equation}
    \hat{p} = \text{softmax}(l)
\end{equation}
The candidate with the highest probability is then chosen as the proper answer.
\section{Experiments}\label{sec:Experiments}
In this section, we conduct evaluations of our model using the ViVQA dataset~\cite{tran-etal-2021-vivqa}. Additionally, we compare our model's performance with that of previous ViVQA models that have achieved state-of-the-art results on the same dataset.

By formulating the VQA task as classification, we employ Cross-Entropy Loss as the loss function and assess our model's performance on the ViVQA dataset, utilizing accuracy as the main metric.

\subsection{Dataset}
As we mentioned in section \ref{vivqadataset}, the ViVQA dataset is the subject of our investigations. A training set of 11,999 Q-A pairs and a test set of 3,001 Q-A pairs comprise the dataset. The detail of the dataset is shown in Table \ref{tab1}.

\begin{table}[h]
\centering
\begin{tabular}{l l l}
\toprule
\textbf{} & \textbf{Train} & \textbf{Test}\\
\midrule
No. Samples    & 11999 &     3001\\
Longest Question Length    & 26 &     24\\
Longest Answer Length     & 4 &     4\\
Average Question Length      & 9.50 &     9.58\\
Average Answer Length    & 1.78 &     1.78\\
\bottomrule
\end{tabular}
\caption{The detail of ViVQA dataset.}\label{tab1}%
\end{table}

\subsection{Evaluation Metrics}
In formulating the ViVQA as a classification problem, we assess the model's effectiveness using key metrics: F1 score, Precision, Recall, and Accuracy. In alignment with our objective to compare against the baseline in our experiment, we prioritize accuracy over other metrics. This choice stems from the fact that all baseline models are evaluated based on accuracy, with some not utilizing additional metrics. Thus, in our experimental endeavor to evaluate the influencing factors of the model, accuracy will serve as the primary metric for a more straightforward analysis.
\subsubsection{Accuracy}
Accuracy or exact match is defined as the ratio of the number of correct predictions to the total number of ground truth answers. Let \(N\) represent the total number of ground truth answers. The accuracy is calculated as follows:
\begin{equation}
    Acuracy = \frac{1}{N}\sum_{i=1}^{N} a_i
\end{equation}
\begin{equation}
    a_i = 
        \begin{cases}
      1, & \text{if}\ \hat{y_i} = y_i \\
      0, & \text{otherwise}
    \end{cases}
\end{equation}
where \(\hat{y_i}\) is the predicted answer, and $y_i$ is the ground truth answer for question \(i\).

\subsubsection{Recall}
This metric, relying on tokenized versions, measures the proportion of correctly identified tokens in the ground truth answer that are also present in the predicted answer. Let $P_i$ represent the tokens in the predicted answer and $GT_i$ denote the tokens in the ground truth answer for the question \(i\)-th. The recall for the question $i$-th is determined as follows:
\begin{equation}
    r_i = \frac{P_i \cap GT_i}{GT_i}
    \label{eq21}
\end{equation}
Let \(N\) represent the total number of ground truth answers, so the overall recall is computed as:
\begin{equation}
    Recall = \frac{1}{N}\sum_{i=1}^{N} r_i
    \label{eq22}
\end{equation}
\subsubsection{Precision}
Precision, in the context of tokenized forms, evaluates how many of the predicted tokens match those in the ground truth answer. Let $P_i$ represent the tokens in the predicted answer and $GT_i$ denote the tokens in the ground truth answer for the question \(i\)-th. The precision for the question \(i\)-th is calculated as follows:
\begin{equation}
    p_i = \frac{P_i \cap GT_i}{P_i}
    \label{eq23}
\end{equation}
Let \(N\) represent the total number of ground truth answers, so the overall recall is computed as:
\begin{equation}
    Precision = \frac{1}{N}\sum_{i=1}^{N} p_i
    \label{eq24}
\end{equation}
\subsubsection{F1 score}
The F1 score is essentially computed from both recall \eqref{eq21} and precision \eqref{eq23} of each individual question. The F1 score for the question \(i\)-th is calculated as follows:
\begin{equation}
    f_i = 
        \begin{cases}
      0, & \text{if}\ p_i = 0,~r_i = 0 \\
      \frac{2 \times p_i \times r_i}{p_i + r_i}, & \text{otherwise}
    \end{cases}
\end{equation}
Let \(N\) represent the total number of ground truth answers, so the overall F1 score is computed as:
\begin{equation}
    F1 = \frac{1}{N}\sum_{i=1}^{N} f_i
\end{equation}

\subsection{Experimental Settings}
In our experiments, the hyperparameters presented in Table \ref{tab2} were utilized. The training process spanned 20 epochs with a batch size of 65, a drop path rate of 0.3, a learning rate of 3e-5, weight decay set at 0.01, and a warmup ratio of 0.1. The input image resolution was 224 x 224. We employed AdamW~\cite{adamw} as the optimizer for updating parameters. The model training took place on a computing system equipped with NVIDIA TESLA P100-16GB GPUs and 32GB of RAM.

\begin{table}[h]
\centering
\begin{tabular}{@{}ll@{}}
\toprule
\textbf{Hyperparameters} & \textbf{Value}\\
\midrule
Epochs    & 20\\
Encoder layers    & 6\\
Encoder attention heads     & 6\\
Batch size      &65\\
AdamW $\epsilon$    & 1e-8\\
AdamW $\beta$       & (0.9, 0.999)\\
Weight decay        & 0.01\\
Learning rate       & 3e-5\\
Learning rate scheduler type    & Cosine\\
Warmup ratio        & 0.1\\
Drop path     & 0.3\\
Dropout         & \xmark\\
\bottomrule
\end{tabular}
\caption{Hyperparameters for training process.}\label{tab2}%
\end{table}

Remarkably, within the multi-modal fusion module, a crucial hyperparameter during training is the drop path rate, also known as stochastic depth~\cite{droppath}. The role of drop path is akin to dropout~\cite{dropout}, a regularization technique employed to counteract overfitting in neural networks. Both methods introduce randomness during training, yet they operate at different levels of network architecture. Drop path randomly omits individual samples within a batch during training, thereby impacting the depth of the network's computational graph. In contrast, dropout randomly eliminates units within layers, thereby altering the connectivity between layers. Drop path selectively bypasses specific connections within the neural network for each sample independently, while dropout affects all units within a layer. By randomly omitting paths (connections) during training, drop path encourages the network to acquire more robust features by discouraging reliance on specific paths. This diversification of learned features aids in mitigating overfitting by facilitating a broader exploration of the network's parameter space.
\subsubsection{Training Strategy}
In our paper, we adopted a systematic approach to training our model to ensure optimal performance. Initially, we partitioned our dataset into a training set and a validation set in an 80:20 split. This division allowed us to fine-tune and determine the most effective parameters for our model, with a particular emphasis on identifying the most suitable hyperparameters, including the drop path rate, the number of encoder layers, and the count of encoder attention heads. For the rest of the hyperparameters, we closely followed the training settings used for fine-tuning the BEiT-3 model~\cite{Wang_2023_CVPR} on the VQAv2 dataset~\cite{goyal2017making}.

To further refine our training process and ascertain the ideal number of training epochs necessary for model convergence, we employed a 5-fold cross-validation strategy on the training set. This approach enabled us to determine that our model reached convergence after an average of 20 epochs across all folds. Armed with this insight, we proceeded to train our model on the complete training set for the identified duration of 20 epochs.

After selecting the optimal hyperparameters, we tested each model ten times using ten different random seeds. This approach allowed us to determine the statistical significance between any two models using a two-sample t-test.

\subsection{Experimental Results}
\subsubsection{Impact of Fusion Operation Types on the model}
In the initial phase of our study, we conducted an exploration to identify the optimal fusion function $\mathcal{F}$, as defined in equation \eqref{eq9}, for our model. This investigation encompassed various strategies, including element-wise multiplication, element-wise addition, and concatenation, applied to integrate both global and local image features. Despite the simplicity of the fusion strategies employed, they demonstrated noteworthy effectiveness in enhancing the overall performance of our model.

Table \ref{tab3} presents the outcomes of our experimentation. For the element-wise operations of multiplication and addition, we have features of visual $\in \mathbb{R}^{32 \times 768}$, and for concatenation, we have features in $\mathbb{R}^{64 \times 768}$. It is evident that the concatenation operation outperforms the other operations, achieving the highest level of performance. Consequently, we have chosen concatenation as the fusion function $\mathcal{F}$ for subsequent experiments.

\begin{figure}[h!]
    \centering
    \includegraphics[width = 0.8\textwidth]{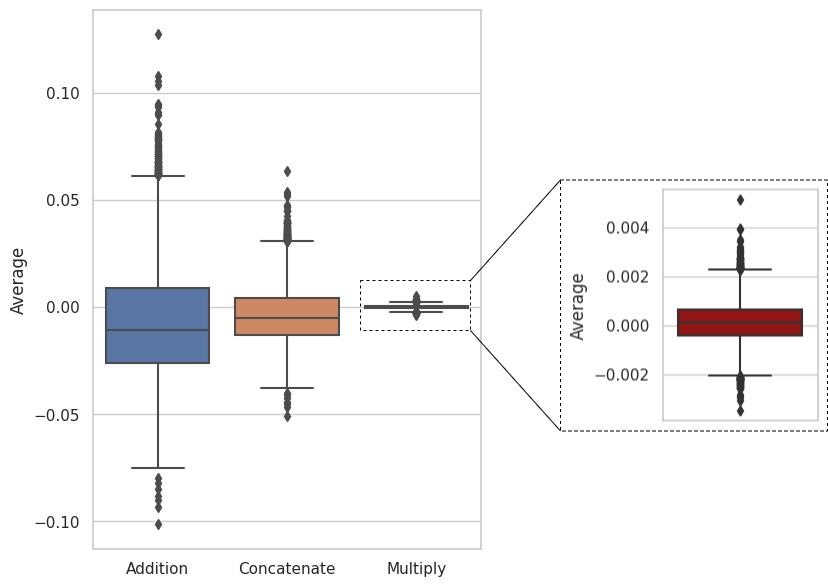}
    \caption{Boxplots illustrate the spread of mean values within the visual features for each operation.}
    \label{fig:boxplot}
\end{figure}

It is noteworthy that the outcome of the multiplication operation is markedly lower compared to concatenation and addition. Our investigation revealed that the diminished performance can be attributed to the multiplication of two features (local $\in \mathbb{R}^{32 \times 768}$ and global $\in \mathbb{R}^{32 \times 768}$) resulting in a new matrix with dimensions $\in \mathbb{R}^{32 \times 768}$ that exhibits a high degree of sparsity. The boxplots in Figure \ref{fig:boxplot} indicate that, on average, the elements in the matrices after addition and concatenation operations demonstrate elevated levels and substantial spreads. In contrast, the multiplication operation yields a sparse matrix characterized by elements that tend to approach zero closely, displaying a narrow range of spreads. This pattern suggests that the sparse matrix resulting from multiplication leads to information loss in the visual representation, thereby accounting for its diminished performance.

\begin{table}[h]
\centering
\begin{tabular}{@{}lr@{}}
\toprule
\textbf{Operation (\(\mathcal{F}\))} & \textbf{Accuracy ($\%$)} $\uparrow$\\
\midrule
Element-wise Multiplication    & 40.15\\
Element-wise Addition    & 69.78\\
\textbf{Concatenation}     & \textbf{71.04}\\
\bottomrule
\end{tabular}
\caption{The effects of fusion operation. $\uparrow$ indicates that higher values are better.}\label{tab3}%
\end{table}

\subsubsection{Impact of CNNs on the model}
In the subsequent phase, we conducted a meticulous investigation into the influence of various CNN architectures on the efficacy of our model. The choice of a CNN architecture plays a pivotal role in shaping our model's performance. We comprehensively evaluated diverse CNN models, including ResNet~\cite{resnet} and VGGNet~\cite{vgg}, alongside the EfficientNet family architecture. This assessment was rooted in accuracy metrics and an examination of the parameter count for each model.

Table \ref{tab4} presents a synthesis of accuracy and parameter count data for each CNN model. The findings reveal that the EfficientNet family not only surpasses other models in terms of accuracy but also, with a moderate parameter count, exhibits superior performance compared to the conventional ResNet and VGG architectures. As observed, VGG achieves an accuracy of over 70$\%$, but it requires hundreds of millions of parameters. In contrast, EfficientNet achieves comparable accuracy with only tens of millions of parameters. We selected the large-scale model (EfficientNet-B7) as the optimal choice, offering elevated accuracy with reasonable model complexity for our model.

\begin{table}[h]
\centering
\begin{tabular}{@{}lrr@{}}
\toprule
\textbf{CNN Models} & \textbf{Accuracy ($\%$)} $\uparrow$ & \textbf{Params}\\
\midrule
Resnet-18    & 68.74   &11.7 M\\
Resnet-34    & 68.61   &21.8 M\\
Resnet-50    & 69.44   &25.6 M\\
Resnet-101    & 69.11   &44.5 M\\
Resnet-152    & 69.94   &60.2 M\\
VGG-11    & 70.48   &132.9 M\\
VGG-13    & 70.44   &133.0 M\\
VGG-16    & 69.37   &138.4 M\\
VGG-19    & 69.81   &143.7 M\\
Efficientnet-B0     & 69.61   &5.3 M\\
Efficientnet-B1     & 68.78   &7.8 M\\
Efficientnet-B2     & 69.38   &9.1 M\\
Efficientnet-B3     & 69.71   &12.2 M\\
Efficientnet-B4     & 69.78   &19.3 M\\
Efficientnet-B5     & 70.64   &30.4 M\\
Efficientnet-B6     & 70.24   &43.0 M\\
\textbf{Efficientnet-B7}     & \textbf{71.04}   &\textbf{66.3 M}\\
\bottomrule
\end{tabular}
\caption{Influence of CNN structures on model performance. $\uparrow$ indicates that higher values are better.}\label{tab4}%
\end{table}

\subsubsection{Comparison with other Methods}
Due to the limited research on VQA tasks in Vietnamese, we benchmarked our model against several existing approaches that have demonstrated promising results in this domain.  Specifically, we compared our model to strategies proposed by other methods. LSTM-based approaches \cite{lstm} and pre-trained word embeddings (PhoW2Vec \cite{tuan-nguyen-etal-2020-pilot}) proved ineffective in handling the complexities of VQA. Conversely, methods based on transformers demonstrated excellent performance in addressing this challenging task.

\begin{table}[h]
\centering
\resizebox{\textwidth}{!}{
\begin{tabular}{@{}lcrccc@{}}
\toprule
\textbf{Models} & \textbf{Accuracy ($\%$)} $\uparrow$ & \textbf{Improvement} $\uparrow$ ($\%$) & \textbf{Precision ($\%$)} $\uparrow$ & \textbf{Recall ($\%$)} $\uparrow$ & \textbf{F1-score ($\%$)} $\uparrow$\\
\midrule
LTSM + PhoW2Vec~\cite{tran-etal-2021-vivqa}    & 33.85 & 37.19 (*)& -&  -& -\\
Bi-LTSM + PhoW2Vec~\cite{tran-etal-2021-vivqa}     & 33.97 & 37.07 (*)& -&  -& -\\
HieCoAtt + PhoW2Vec~\cite{tran-etal-2021-vivqa}    & 34.96 & 36.08 (*)& -&  -& -\\
PAT~\cite{pat}                    & 60.55 & 10.49 (*)& -&  -& -\\
MCA~\cite{CMSA}                    & 60.76 & 10.28 (*)& -&  -& -\\
BARTPhoBEiT~\cite{bartphobeit}& 68.58 & 2.46 (*)& 69.31&  68.58& 67.77\\
\textbf{Our model}               & \textbf{71.04} & -& \textbf{76.34}&  \textbf{76.29}& \textbf{76.16}\\
\bottomrule
\end{tabular}
}
\caption{Our model versus former baseline methods. We computed the percentage improvement in accuracy of our model compared to other baselines. (*) indicates statistically significant ($p < 0.05$) improvements. $\uparrow$ indicates that higher values are better.}\label{tab5}%
\end{table}

The comparative results are presented in Table \ref{tab5}. Our model achieves a remarkable accuracy of 71.04\% on the test set, surpassing all other methods by a significant margin. This represents a substantial improvement of 37.19\% over the least accurate model (LSTM + PhoW2Vec) and a notable gain of 2.46\% over the second-best approach (BARTPhoBEiT). These results unequivocally demonstrate the superiority of our proposed method in addressing ViVQA.

Overall, many previous baselines lack information about local features, with the exception of MCA. The effectiveness of incorporating both local and global features has proven to be crucial for the success of previous literature. We will conduct an ablation study in the next section to demonstrate their influence. Although MCA considers both local and global features in image extraction and achieves commendable accuracy, it utilizes ViT, which is trained solely on image datasets. In contrast, our method leverages BLIP-2, which is pre-trained on both vision and language data. This vision-language pre-training enables BLIP-2 to capture more comprehensive and contextually rich information than that obtained from ViT.

With promising results achieved on the ViVQA dataset, our model has demonstrated effectiveness in the VQA task, particularly in its ability to extract, represent, and comprehend the relationship between images and text. To ensure the model effectively understands Vietnamese, we propose leveraging not only a powerful pre-trained model like BARTPho but also integrating a multi-modal information synthesis module like BEiT-3 to create a joint representation containing information from both text and images. Additionally, our model not only relies on the existing knowledge encoded in the pre-trained BARTPho model for Vietnamese, but we also refine it to suit multi-modal tasks between Vietnamese images and texts, thereby establishing a foundation for similar multi-modal studies in Vietnamese.


\subsection{Ablation Study}
We conducted an ablation study to thoroughly investigate the contributions of BLIP-2 and EfficientNet to the overall outcome. As depicted in Table \ref{tab6}, models relying solely on BLIP-2 or EfficientNet for extracting image features exhibited lower accuracy. Whilst combining both methods led to improved performance, surpassing the results achieved with a singular approach.
\begin{table}[h]
\centering
\resizebox{\textwidth}{!}{
\begin{tabular}{@{}lcrccc@{}}
\toprule
\textbf{Visual Extractor} & \textbf{Accuracy ($\%$)} $\uparrow$& \textbf{Improvement} $\uparrow$ ($\%$) & \textbf{Precision ($\%$)} $\uparrow$& \textbf{Recall ($\%$)} $\uparrow$& \textbf{F1-score ($\%$)} $\uparrow$\\
\midrule
EfficientNet-B7    & 53.95 & 17.09 (*)& 61.92& 61.83& 61.67\\
BLIP-2    & 69.61 & 1.43 (*)& 74.82& 74.83& 74.66\\
\textbf{BLIP-2 + EfficientNet-B7}     & \textbf{71.04} & -& \textbf{76.34}& \textbf{76.29}& \textbf{76.16}\\
\bottomrule
\end{tabular}
}
\caption{Ablation study for visual extractor. We computed the percentage improvement in accuracy of the combination of two visual extractors compared to using only one (*) indicates statistically significant ($p < 0.05$) improvements. $\uparrow$ indicates that higher values are better.}\label{tab6}%
\end{table}

\begin{figure}[h!]
    \centering
    \includegraphics[width = 0.6\textwidth]{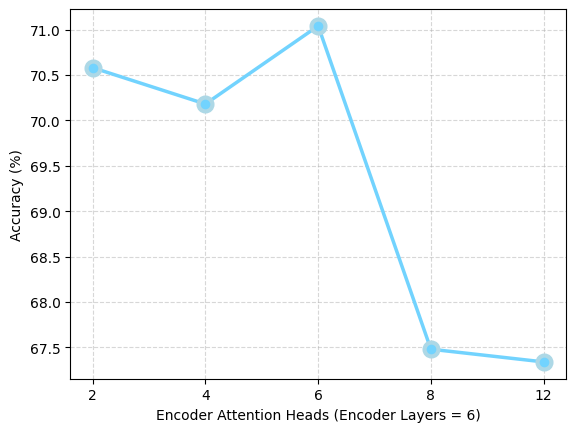}
    \caption{Ablation study for encoder attention heads.}
    \label{fig:attn_heads}
\end{figure}

Furthermore, we systematically selected the number of encoder attention heads and encoder layers in the multi-modal fusion module. In our system, this module plays a pivotal role in the model’s performance for combining information from the visual and question modalities. Initially, we fixed the number of encoder layers at $6$ and varied the number of encoder attention heads. The results, as illustrated in Figure \ref{fig:attn_heads}, highlight the optimal performance of our model with 6 encoder attention heads.

\begin{figure}[h!]
    \centering
    \includegraphics[width = 0.6\textwidth]{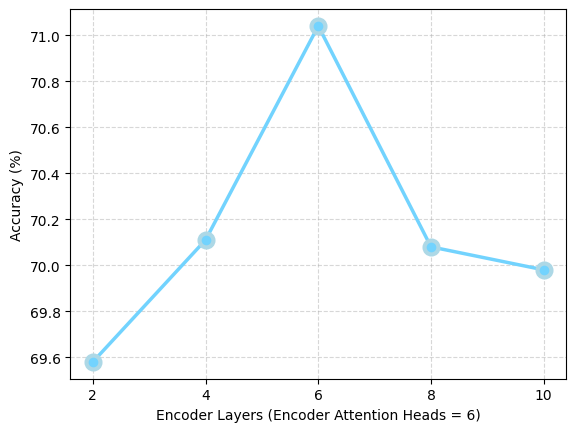}
    \caption{Ablation study for encoder layers.}
    \label{fig:encoder_layers}
\end{figure}

Subsequently, maintaining 6 encoder attention heads constant, we adjusted the number of encoder layers. The outcomes, showcased in Figure \ref{fig:encoder_layers}, underscore that the optimal number of layers for the encoder is 6.

After obtaining the ultimate components of our model, we investigated the impact of freezing the visual extractor on model performance. Due to limited infrastructure in the experimental environment, we only used EfficientNet-B7 from our image embedding module to conduct the experiment. The results, depicted in Table \ref{tab7}, indicate that freezing the EfficientNet-B7 led to an accuracy of 71.04$\%$ within 20 epochs, whereas the visual extractor is not frozen, the accuracy is slightly lower at 70.44$\%$, with a significantly longer training time of 935 seconds. From these findings, we conclude that freezing the visual extractor can enhance efficiency without compromising accuracy, as well as saving a lot of training time, rendering it a viable strategy for training deep learning models in similar contexts.

\begin{table}[h]
\centering
\resizebox{\textwidth}{!}{
\begin{tabular}{@{}lcccc@{}}
\toprule
\textbf{Visual Extractor} & \textbf{Freeze}& \textbf{Accuracy ($\%$)} $\uparrow$ & \textbf{Epoch} & \textbf{Training Time (s) $\downarrow$}\\
\midrule
\multirow{2}{*}{EfficientNet-B7}  & \cmark & \textbf{71.04}& 20 &225\\
                & \xmark & 70.44& 20& 935  \\
\bottomrule
\end{tabular}
}
\caption{Ablation study for freezing the visual extractor. $\uparrow$ indicates that higher values are better, while $\downarrow$ indicates that lower values are better.}\label{tab7}%
\end{table}

\subsection{Discussion}
\subsubsection{Dataset Challenges}
\begin{figure}[ht!]
\centering
\begin{subfigure}[t]{0.3\textwidth}
    \includegraphics[width=\textwidth]{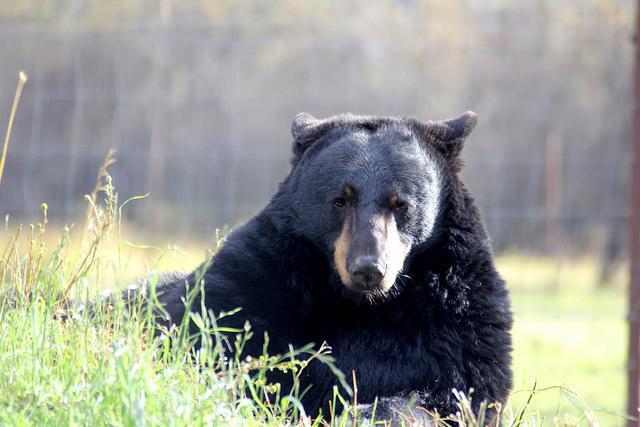}
    \caption{
        Example 1 \\
        \textbf{Question}: \textvietnamese{Màu của gấu là gì?} (What is the color of the bear?)\\
        \textbf{Ground Truth}: \textvietnamese{Màu nâu} (Brown)
    }
    \label{fig:ex1}
\end{subfigure}
\hfill
\begin{subfigure}[t]{0.3\textwidth}
    \includegraphics[width=\textwidth]{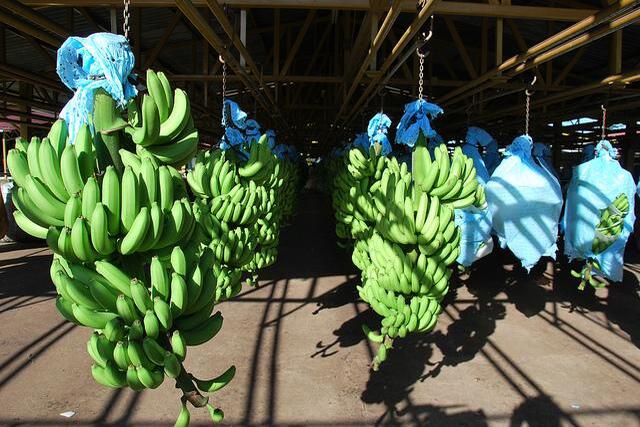}
    \caption{
        Example 2 \\
        \textbf{Question}: \textvietnamese{Một số chuối xanh treo ở đâu?} (Where are some green bananas?) \\
        \textbf{Ground Truth}: \textvietnamese{Tòa nhà} (Building)
    }
    \label{fig:ex2}
\end{subfigure}
\hfill
\begin{subfigure}[t]{0.3\textwidth}
    \includegraphics[width=\textwidth]{Fig/examples/336320.jpg}
    \caption{
        Example 3 \\
        \textbf{Question}: \textvietnamese{Những gì được phủ lên với một số thành phần ngon?} (What's topped some delicious ingredients?)\\
        \textbf{Ground Truth}: \textvietnamese{Con chó} (Dog)
    }
    \label{fig:ex3}
\end{subfigure}
\hfill
\begin{subfigure}[t]{0.3\textwidth}
    \includegraphics[width=\textwidth]{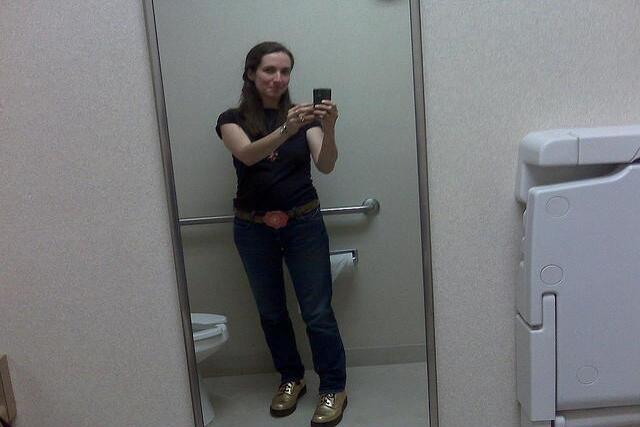}
    \caption{
        Example 4 \\
        \textbf{Question}: \textvietnamese{Cô gái đang chụp ảnh selfie ở đâu?} (Where is the girl taking a selfie?)\\
       \textbf{Ground Truth}: \textvietnamese{Chuồng} (Cage)
    }
    \label{fig:ex3}
\end{subfigure}
\hfill
\begin{subfigure}[t]{0.3\textwidth}
    \includegraphics[width=\textwidth]{Fig/examples/576052.jpg}
    \caption{
        Example 5 \\
        \textbf{Question}: \textvietnamese{Người đó đã làm gì với một con ngựa?} (What did that person do to a horse?)\\
       \textbf{Ground Truth}: \textvietnamese{Núi} (Mountain)
    }
    \label{fig:ex3}
\end{subfigure}
\hfill
\begin{subfigure}[t]{0.3\textwidth}
    \includegraphics[width=\textwidth]{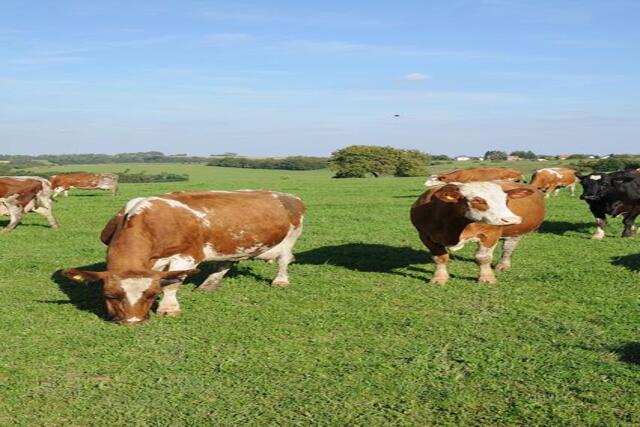}
    \caption{
        Example 6 \\
        \textbf{Question}: \textvietnamese{Màu của con bò là gì?} (What is the color of the cow?)\\
      \textbf{Ground Truth}: \textvietnamese{Màu trắng} (White)
    }
    \label{fig:ex3}
\end{subfigure}
\hfill
\caption{A few instances of low-quality examples.}
\label{fig:low_quality_examples}
\end{figure}

Despite the comprehensive validation of the dataset by professionals, certain samples exhibit ambiguity, grammatical issues in ViVQA questions, or a lack of relevance between questions and answers. This highlights the challenge posed by the dataset's overall low quality, a significant hurdle for VQA tasks in Vietnamese, primarily due to the scarcity of high-quality datasets. As depicted in Figure \ref{fig:low_quality_examples}, we illustrate instances where the ground truth answers are inappropriate, unclear, or unrelated to the corresponding questions. Particularly, discrepancies arise, such as the misidentification of the bear's color as brown instead of black, or the color of the cows being described as brown while the ground truth indicates white. Moreover, descriptions indicating a sandwich topped with quality ingredients are contradicted by an irrelevant answer, $`$Dog'. Similarly, instances where the depicted scene clearly shows a man riding a horse but is misinterpreted as a $`$mountain', and a girl taking a selfie, potentially in a restroom, yet labeled as a $`$Cage', further underscore the poor quality of the dataset in this study. These findings highlight the poor quality of the ViVQA dataset in this study, presenting additional challenges for researchers interested in improving dataset quality.


\subsubsection{Experimental Analysis}
\begin{figure}[ht!]
\centering
\begin{subfigure}[t]{0.3\textwidth}
    \includegraphics[width=\textwidth]{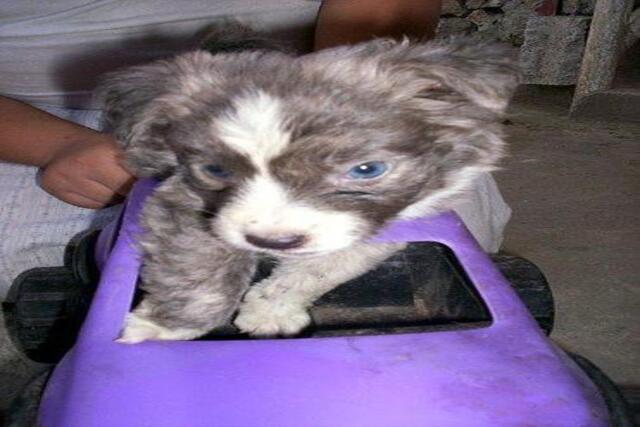}
    \caption{
        Example 1 \\
        \textbf{Question}: \textvietnamese{Con chó nhỏ ở đâu?} (Where is the little dog?)\\
        \textbf{Ground Truth}: \textcolor{green!70!black}{\textvietnamese{Xe ô tô} (Car)}\\
        \textbf{BLIP-2}: \textcolor{red}{\textvietnamese{Xe tải} (Truck)}\\
        \textbf{EfficientNet}: \textcolor{red}{\textvietnamese{Vali} (Suitcase)}\\
        \textbf{Our Model}: \textcolor{green!70!black}{\textvietnamese{Xe ô tô} (Car)}
    }
    \label{fig:ex1}
\end{subfigure}
\hfill
\begin{subfigure}[t]{0.3\textwidth}
    \includegraphics[width=\textwidth]{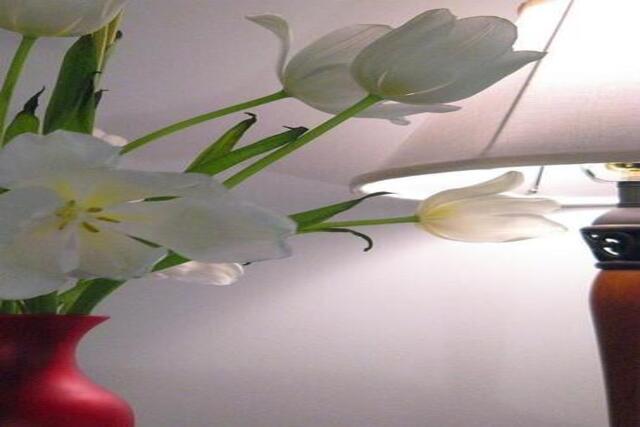}
    \caption{
        Example 2 \\
        \textbf{Question}: \textvietnamese{Màu của chiếc bình là gì?} (What is the color of the vase?) \\
        \textbf{Ground Truth}: \textcolor{green!70!black}{\textvietnamese{Màu đỏ} (red)}\\
        \textbf{BLIP-2}: \textcolor{red}{\textvietnamese{Màu trắng} (white)}\\
        \textbf{EfficientNet}: \textcolor{green!70!black}{\textvietnamese{Màu đỏ} (red)}\\
        \textbf{Our Model}: \textcolor{green!70!black}{\textvietnamese{Màu đỏ} (red)}
    }
    \label{fig:ex2}
\end{subfigure}
\hfill
\begin{subfigure}[t]{0.3\textwidth}
    \includegraphics[width=\textwidth]{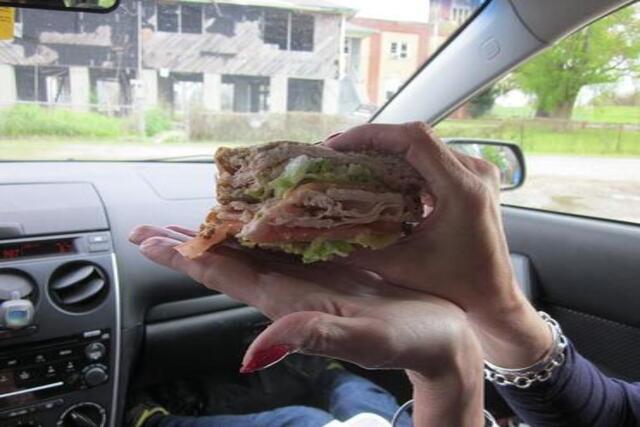}
    \caption{
        Example 3 \\
        \textbf{Question}: \textvietnamese{Người phụ nữ đang giữ bánh sandwich ở đâu?} (Where is the woman holding the sandwich?)\\
        \textbf{Ground Truth}: \textcolor{green!70!black}{\textvietnamese{Xe ô tô} (Car)}\\
        \textbf{BLIP-2}: \textcolor{green!70!black}{\textvietnamese{Xe ô tô} (Car)}\\
        \textbf{EfficientNet}: \textcolor{red}{\textvietnamese{Phòng bếp} (Kitchen)}\\
        \textbf{Our Model}: \textcolor{green!70!black}{\textvietnamese{Xe ô tô} (Car)}
    }
    \label{fig:ex3}
\end{subfigure}
\hfill
\caption{Examining experimental scenarios to illuminate model performance.}
\label{fig:exper_examples}
\end{figure}

To start, we present insightful examples to analyze the performance of our model. Through experimental analysis, strengths and weaknesses become readily apparent in practical scenarios. The specifics of these instances are illustrated in Figure \ref{fig:exper_examples}. 

In Figure \ref{fig:exper_examples}-(a), both BLIP-2 and EfficientNet made individual incorrect predictions, suggesting that each model lacks comprehensive information from the image. BLIP-2, emphasizing the global context of the image, may overlook local details, while EfficientNet may struggle to capture broader contextual information. 

In Figure \ref{fig:exper_examples}-(b), the model solely relying on BLIP-2 for visual extraction erroneously predicted the color of the vase as white, likely due to BLIP-2's tendency to prioritize global features, where much of the image appears white. Consequently, EfficientNet provided a correct answer by considering local details and determining the color of the vase to be red. 

In Figure \ref{fig:exper_examples}-(c), the model utilizing only EfficientNet for visual extraction predicted that the location where a woman is holding a sandwich is the kitchen. This is attributed to EfficientNet's capability to focus on intricate details such as the sandwich and the dashboard of a car, which might resemble oven buttons. Conversely, BLIP-2, taking into account the global context, accurately identified the location as a car. 

Through experimental evaluation, it became evident that our model, along with its constituent components BLIP-2 and EfficientNet, exhibits both strengths and weaknesses. BLIP-2 excels in capturing the global context of images but may overlook finer local details, while EfficientNet focuses on intricate local features but may struggle with broader contextual information. These insights highlight the importance of considering both global and local features in model development for visual understanding tasks.

\begin{figure}[ht!]
\centering
\begin{subfigure}[t]{0.3\textwidth}
    \includegraphics[width=\textwidth]{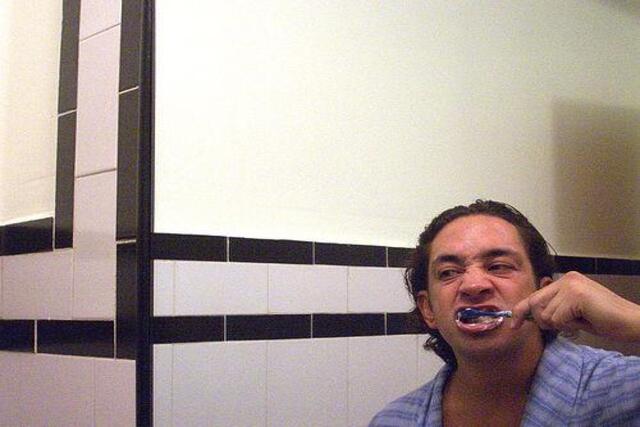}
    \caption{
        Example 1 \\
        \textbf{Question}: \textvietnamese{Người đàn ông đang đánh răng ở đâu?} (Where is the man brushing his teeth?)\\
        \textbf{Ground Truth}: \textcolor{green!70!black}{\textvietnamese{Phòng} (Room)}\\
        \textbf{BLIP-2}: \textcolor{red}{\textvietnamese{Phòng tắm} (Bathroom)}\\
        \textbf{EfficientNet}: \textcolor{red}{\textvietnamese{Phòng tắm} (Bathroom)}\\
        \textbf{Our Model}: \textcolor{red}{\textvietnamese{Phòng tắm} (Bathroom)}
    }
    \label{fig:ex1}
\end{subfigure}
\hfill
\begin{subfigure}[t]{0.3\textwidth}
    \includegraphics[width=\textwidth]{Fig/examples/342295.jpg}
    \caption{
        Example 2 \\
        \textbf{Question}: \textvietnamese{Tác phẩm nghệ thuật được đóng khung và lưu trữ ở đâu?} (Where is the artwork framed and stored?) \\
        \textbf{Ground Truth}: \textcolor{green!70!black}{\textvietnamese{Bồn tắm} (Bathtub)}\\
        \textbf{BLIP-2}: \textcolor{red}{\textvietnamese{Phòng tắm} (Bathroom)}\\
        \textbf{EfficientNet}: \textcolor{red}{\textvietnamese{Chậu} (Pot)}\\
        \textbf{Our Model}: \textcolor{red}{\textvietnamese{Phòng tắm} (Bathroom)}
    }
    \label{fig:ex2}
\end{subfigure}
\hfill
\begin{subfigure}[t]{0.3\textwidth}
    \includegraphics[width=\textwidth]{Fig/examples/540253.jpg}
    \caption{
        Example 3 \\
        \textbf{Question}: \textvietnamese{Những gì chứa đầy những sinh viên nhìn chằm chằm và xung quanh một bảng đầy máy tính xách tay mở?} (What's filled with students staring at and around a table full of open laptops?)\\
        \textbf{Ground Truth}: \textcolor{green!70!black}{\textvietnamese{Lớp học} (Class)}\\
        \textbf{BLIP-2}: \textcolor{red}{\textvietnamese{Phòng} (Room)}\\
        \textbf{EfficientNet}: \textcolor{red}{\textvietnamese{Máy vi tính} (Computer)}\\
        \textbf{Our Model}: \textcolor{red}{\textvietnamese{Phòng} (Room)}
    }
    \label{fig:ex3}
\end{subfigure}
\hfill
\caption{Ambiguity assessment in question-answer datasets.}
\label{fig:exper_examples2}
\end{figure}

Another aspect we wish to address is the ambiguity observed in the answers provided in the dataset utilized for evaluation, as well as the complexity of the linguistic aspects. In Figures \ref{fig:exper_examples2}-(a,b), ambiguity arises between the ground truth answer and the predicted answer. We believe that the predicted answer could potentially be correct regarding the location of the object in the image; specifically, the man and artwork could be situated in the bathroom. Furthermore, in Figure \ref{fig:exper_examples2}-(c), there is an excessive amount of redundant information contained in the question, making it difficult for our model to extract relevant information. Consequently, our model predicts the overall scene (Room) instead of a specific place (Class). This underscores the importance of addressing ambiguity and redundant information in question-answer datasets, as they can significantly impact model performance and interpretation. With a sufficiently robust dataset, we believe we could tackle this challenge, representing the motivation driving our future focus.

\begin{figure}[h!]
    \centering
    \includegraphics[width = 0.5\textwidth]{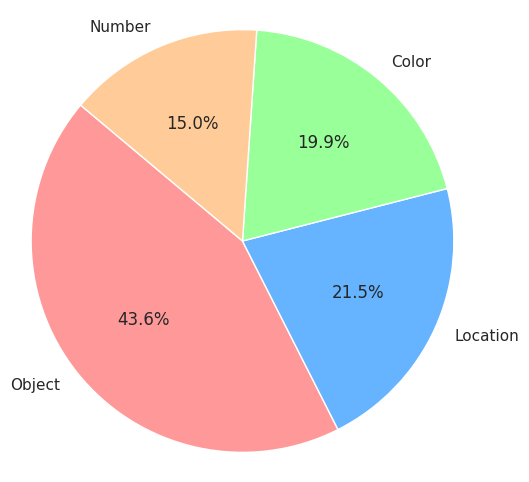}
    \caption{Distribution of incorrectly predicted question types.}
    \label{fig:distribution}
\end{figure}

An additional noteworthy aspect we would like to address pertains to the distribution of incorrectly predicted question types by our model. As depicted in Figure \ref{fig:distribution}, our model inaccurately predicted the question type $`$object' in approximately 43.6$\%$ of cases, which accounts for nearly half of all incorrect predictions. This observation suggests that our model may lack sufficient information regarding the specific details of objects within the image. Consequently, addressing this limitation stands as a motivating factor for our future work.

\subsubsection{Transfer Learning Considerations}
Our architecture is capable of efficiently transferring knowledge from a source domain, where the model undergoes initial training, to a target domain characterized by limited data or a demand for rapid deployment. This capability is underpinned by the flexibility of our design, which facilitates the easy replacement and adaptation of the backbone components.

In the context of linguistic applications, while our primary focus has been on monolingual Vietnamese, the model architecture can be readily adapted to other monolingual languages with minimal modifications to the Text Embedding module presented in section \ref{textembedding}. This adaptability ensures that the linguistic component of our model can handle a diverse array of languages, thereby broadening its applicability. Similarly, in the visual domain, the model's backbones presented in section \ref{imageembedding}  can be replaced with alternative architectures, providing the flexibility to optimize performance based on specific visual tasks or datasets. 

Moreover, the versatility of our model extends beyond typical visual-linguistic tasks to specialized domains, such as Medical Visual Question Answering (MedVQA). Although our current work primarily addresses everyday scenes and objects, the transition to the medical domain is facilitated by the consistent input format of text-image pairs. This consistency aligns seamlessly with our architecture, enabling the model to leverage pre-existing knowledge and adapt to the nuanced requirements of medical imagery and terminology.

The ability to transfer learning effectively across these varied domains underscores the robustness of our architecture. By allowing for the modular exchange of core components, such as the text and visual backbones, our model provides a scalable and adaptable solution that meets the demands of diverse applications. This not only enhances the model’s utility across different linguistic and visual tasks but also ensures that it can be efficiently repurposed for new, domain-specific challenges.
\section{Conclusion and Future Work}\label{sec:Conclusion}
In this study, we introduced a ViVQA system with the key integration of convolutional and base-transformer methods for vision embedding. The convolutional methods capture local features, while the base-transformer methods address global features, resulting in a more comprehensive image representation. Our experiments demonstrated the effectiveness of these combined approaches in enhancing our architecture's performance. However, we also encountered challenges related to the quality of the ViVQA dataset. The dataset's low quality posed significant difficulties, as it included several incorrectly labeled samples, which affected the overall performance. Despite our model's correct predictions from a human-like perspective, these predictions were not reflected accurately in the performance evaluations due to the dataset's inaccuracies. Furthermore, our model's performance is limited when addressing open-ended questions. The current dataset only encompasses four types of questions, and questions that deviate from these predefined types are not handled effectively by our model.

Despite the promising results, we recognize the need to address several challenges in future work. Our plans include improving the quality of the ViVQA dataset by conducting thorough reviews to ensure the accuracy of both questions and answers. We also intend to employ data augmentation techniques to further enrich the dataset. Additionally, as analyzed, our model still faces difficulties with questions concerning objects. To overcome this limitation, we aim to enhance object information by incorporating object detection models into the visual extraction process.

Our objective is to significantly improve the overall performance of the ViVQA system and contribute to the advancement of more efficient and reliable natural language processing in Vietnamese. By addressing these challenges, we hope to develop a more robust and accurate ViVQA system capable of handling a wider range of queries and providing better support for Vietnamese language applications.
\section*{Acknowledgement}
This research is funded by University of Science, VNU-HCM, Vietnam under grant number CNTT 2024--12.

\bibliographystyle{elsarticle-num-names} 
\bibliography{reference.bib}





\end{document}